\documentclass[10pt,leqno]{amsart}
\usepackage{graphicx}
\usepackage{indentfirst,csquotes}
\usepackage{amssymb,amsthm,amsmath}
\usepackage{xcolor,paralist,hyperref,titlesec,fancyhdr,etoolbox}
\usepackage{booktabs,multirow,array,float,placeins}

\titleformat{\section}[display]{\normalfont\huge\bfseries\centering}{\centering\chaptertitlename\thechapter}{10pt}{\Large}
\titlespacing*{\section}{0pt}{0ex}{0ex}
\hypersetup{ colorlinks=true, linkcolor=black, filecolor=black, urlcolor=black }

\begin{document}

\title{MVSegNet: A Lightweight Boundary-Aware Network for Fetal Lateral
Ventricle Segmentation and Atrial Width Estimation in Prenatal Ultrasound}

\author[A. H. Sayem]{Arafat Hossain Sayem}

\date{\today}

\address{Department of Computer Science \& Engineering,
Stamford University Bangladesh, Dhaka 1217, Bangladesh}

\email{sayem.cse72@gmail.com}

\maketitle

\let\thefootnote\relax

\begin{abstract}
\textbf{Purpose:} Fetal ventriculomegaly is assessed by measuring the
atrial width of the lateral ventricle in prenatal ultrasound. Accurate
segmentation is essential for this measurement, but acoustic shadowing,
speckle noise, and poor contrast make it difficult.
\textbf{Methods:} We developed MVSegNet, a lightweight encoder-decoder
network combining multi-scale feature extraction and boundary-aware
refinement. The model was trained and evaluated on 584 expert-annotated
transventricular ultrasound frames using a 70/15/15 split. Performance
was compared against six segmentation baselines using overlap, boundary,
and measurement metrics.
\textbf{Results:} MVSegNet achieved a Dice score of 80.79\%, IoU of
68.47\%, Hausdorff distance of 4.07 mm, and atrial width mean absolute
error of 3.40 mm. The model contains 2.31 million parameters and runs
at 165.6 frames per second on an NVIDIA T4 GPU.
\textbf{Conclusions:} MVSegNet outperformed all evaluated baselines on
boundary and measurement metrics while maintaining low computational
cost, supporting its use in automated fetal ultrasound analysis.
\end{abstract}

\bigskip

\section*{1. Introduction}

Fetal ventriculomegaly is one of the most common central nervous system
findings in prenatal ultrasound. It is assessed by measuring the atrial
width of the lateral ventricle on the transventricular plane. This
measurement depends on accurate identification of the ventricular
boundary, so reliable segmentation is important for consistent clinical
evaluation.

In routine practice, atrial width is measured manually using electronic
calipers. This process is subject to inter-observer and intra-observer
variability. The task is further complicated by acoustic shadowing, low
contrast, speckle noise, and anatomical variation, all of which can
degrade the ventricle boundary in ultrasound images.

Encoder-decoder networks have become the standard approach for medical
image segmentation. Ronneberger et al.~\cite{ronneberger2015unet}
established this design using hierarchical feature extraction with skip
connections. Zhou et al.~\cite{zhou2018unetplusplus} extended this by
redesigning skip pathways for richer feature aggregation, and Oktay et
al.~\cite{oktay2018attentionunet} added attention gates to suppress
irrelevant encoder responses. Chen et al.~\cite{chen2018deeplabv3plus}
used atrous separable convolutions with spatial pyramid pooling to
capture wider context. More recently, foundation models such as
Kirillov et al.~\cite{kirillov2023sam} and Ma et al.~\cite{ma2024medsam}
have extended segmentation to broader settings through large-scale
pretraining.

Despite these advances, fetal lateral ventricle segmentation remains
difficult. The ventricle varies in size and shape across gestational
ages, and its boundary is often weak or obscured. Models that score
well on overlap metrics do not always produce reliable contours in hard
cases. Lightweight models are also important because practical
deployment often involves limited hardware. Howard et
al.~\cite{howard2019mobilenetv3} offers a good balance between
efficiency and accuracy and serves as the shared encoder backbone for
both MobileNet-UNet and the proposed MVSegNet in this study.

We propose MVSegNet, a lightweight encoder-decoder network for fetal
lateral ventricle segmentation. The model combines multi-scale feature
extraction and boundary-aware refinement to improve contour quality
while keeping computational cost low. We evaluate it against six
baselines on 584 expert-annotated transventricular frames from a public
benchmark \cite{alzubaidi2023dataset}.

The main contributions are:
\begin{enumerate}
    \item MVSegNet, a lightweight segmentation network that improves
    boundary quality and measurement accuracy while remaining
    computationally efficient.
    \item A comparison against six baselines showing MVSegNet achieves
    the best overall performance on Dice, IoU, boundary metrics, and
    inference speed.
    \item Evidence that a compact task-specific model can outperform
    larger general-purpose architectures on this low-resource task.
\end{enumerate}

\section*{2. Materials and Methods}

\subsection*{2.1. Overview}

MVSegNet is a lightweight encoder-decoder network for fetal lateral
ventricle segmentation from transventricular ultrasound images. Its
architecture combines a lightweight pretrained encoder, multi-scale
feature enhancement, attention-guided skip fusion, and boundary-aware
refinement. A schematic overview is shown in Fig.~1.

\begin{figure}[H]
    \centering
    \includegraphics[width=\textwidth]{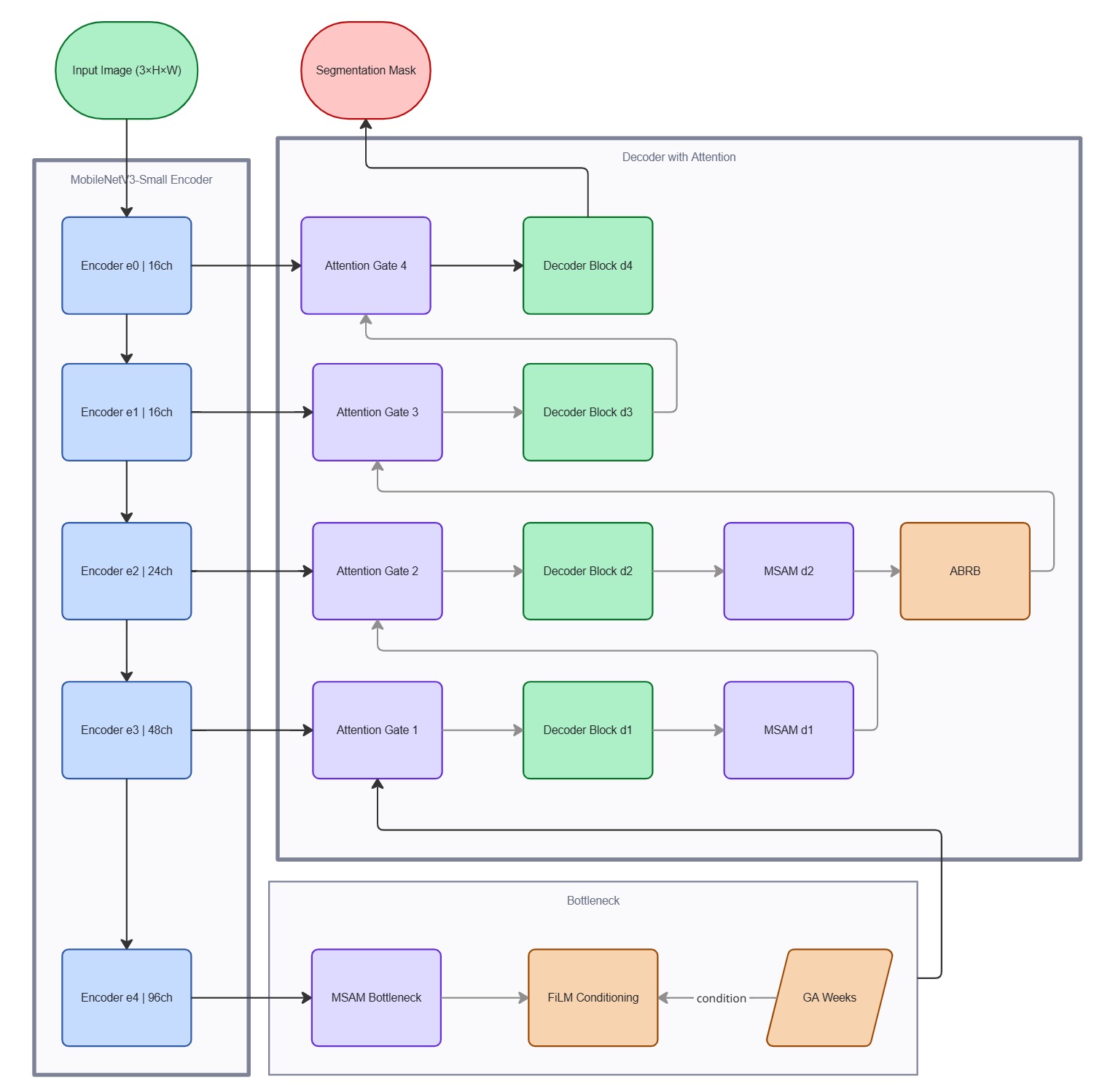}
    \caption{Overview of the MVSegNet architecture. A MobileNetV3-Small
    encoder extracts multi-resolution feature maps, which are decoded
    through four progressive upsampling stages. MSAM is applied at the
    bottleneck and two intermediate decoder stages. Attention gates
    filter encoder skip features before fusion with decoder features.
    ABRB is inserted at the second decoder stage to refine
    boundary-sensitive features. The final output is a lateral ventricle
    segmentation map. Auxiliary segmentation outputs are used during
    training for deep supervision only.}
    \label{fig:architecture}
\end{figure}
\FloatBarrier

\subsection*{2.2. Encoder}

The encoder is based on MobileNetV3-Small \cite{howard2019mobilenetv3},
selected for its low computational cost and efficient depthwise
separable convolution design. The backbone is initialised with
ImageNet-pretrained weights and divided into five sequential stages
that generate multi-resolution feature maps $\mathbf{e}_0$ through
$\mathbf{e}_4$. Shallower features preserve spatial detail and the
deepest features encode high-level contextual information.

\subsection*{2.3. Feature-wise Conditioning with Gestational Age}

MVSegNet applies Feature-wise Linear Modulation
(FiLM) \cite{perez2018film} at the bottleneck stage to condition the
deepest encoder features using gestational age. Let
$\mathbf{e}_4 \in \mathbb{R}^{C \times H \times W}$ denote the
bottleneck feature map and let $g$ denote gestational age in weeks.
A lightweight conditioning branch maps $g$ to two channel-wise
parameter vectors,
$\boldsymbol{\gamma}(g), \boldsymbol{\beta}(g) \in \mathbb{R}^{C}$,
which modulate the bottleneck features as
\begin{equation}
\tilde{\mathbf{e}}_4 = \boldsymbol{\gamma}(g) \odot \mathbf{e}_4
+ \boldsymbol{\beta}(g),
\end{equation}
where $\odot$ denotes channel-wise multiplication broadcast across
spatial dimensions. Two fully connected layers generate the scaling
and shifting coefficients. The FiLM layer is initialised as an
identity transformation. If gestational age is unavailable at
inference time, the FiLM operation is omitted.

\subsection*{2.4. Multi-Scale Ventricle Attention Module}

MVSegNet employs a Multi-Scale Ventricle Attention Module (MSAM) to
capture both local boundary detail and broader anatomical context.
MSAM processes input features through four parallel branches: a
$1\times1$ convolution branch, a $3\times3$ depthwise-separable
branch, a $5\times5$ depthwise-separable branch, and an
average-pooling branch followed by projection. The outputs are
concatenated to form a multi-scale representation, then refined by
channel attention and spatial attention. MSAM is applied at the
bottleneck and at two intermediate decoder stages.

\subsection*{2.5. Attention Gates on Skip Connections}

Attention gates are placed on all encoder-decoder skip connections.
At each decoding stage, the gating signal from the decoder and the
corresponding encoder feature map are combined to produce an attention
coefficient map. This suppresses irrelevant background activations
and retains more task-relevant spatial responses, which is useful in
low-contrast ultrasound images with acoustic artefacts.

\subsection*{2.6. Decoder}

The decoder consists of four progressive upsampling blocks. Each block
upsamples the incoming feature map using bilinear interpolation,
applies a $1\times1$ convolution for channel reduction, concatenates
the attended skip features, then passes the result through two
$3\times3$ convolutional layers with batch normalisation and ReLU
activation. Bilinear upsampling is used instead of transposed
convolution to reduce checkerboard artefacts. Spatial dropout is
applied after each decoder block, with higher rates at coarser stages.

\subsection*{2.7. Adaptive Boundary Refinement Block}

An Adaptive Boundary Refinement Block (ABRB) is placed at the second
decoder stage. ABRB first generates a single-channel edge logit map
from the decoder features. The sigmoid-activated edge response is
concatenated with the original decoder features and processed through
a lightweight boundary-refinement pathway. The refined features are
fused back with the original feature stream through a residual-style
connection.

\subsection*{2.8. Output Heads}

\textbf{Segmentation head.} The final decoder output passes through
two convolutional blocks and a $1\times1$ projection to produce the
final segmentation logits, then upsampled to the input resolution.

\textbf{Auxiliary heads.} Two auxiliary segmentation heads are
attached to intermediate decoder stages for deep supervision during
training. Each consists of a $1\times1$ convolution followed by
bilinear upsampling. These are not used at inference.

\textbf{Boundary output.} ABRB produces an intermediate edge logit
map that supports boundary-aware feature refinement during training.

\subsection*{2.9. Loss Function}

The primary segmentation output is supervised using a combination of
binary cross-entropy and Tversky loss to balance pixel-wise
discrimination and mask-level agreement. Auxiliary supervision is
applied to intermediate outputs to stabilise decoder learning. The
boundary-refinement branch is supervised using edge targets derived
from the ground-truth masks.

\subsection*{2.10. Training Protocol}

All models were implemented in PyTorch and trained on the 584-image
dataset using a 70\%/15\%/15\% split. Images and masks were resized
to $512 \times 512$ pixels. Data augmentation included horizontal
flipping, shift-scale-rotation, elastic deformation, random
brightness-contrast adjustment, Gaussian noise, and blur. Validation
and test images were resized only.

Models were trained for 50 epochs with batch size 8 using the AdamW
optimiser with learning rate $10^{-3}$ and weight decay $10^{-4}$.
A OneCycle learning-rate schedule with cosine annealing was applied.
Mixed-precision training was enabled and model selection was based on
the best validation Dice score. All experiments were conducted on an
NVIDIA Tesla T4 GPU.

\section*{3. Results}

\subsection*{3.1. Dataset and Evaluation Protocol}

All experiments were conducted on the publicly available
transventricular ultrasound dataset released by
\cite{alzubaidi2023dataset}, comprising 584 expert-annotated fetal
transventricular ultrasound frames acquired across second- and
third-trimester gestational ages. The dataset carries no official
train/test partition and was released in fully de-identified form
with no patient-level identifiers, so an image-level split was the
only feasible protocol. We applied a stratified random split with a
fixed seed (42) into training (408 frames, 70\%), validation (88
frames, 15\%), and test (88 frames, 15\%) sets. This split was held
fixed across all experiments.

Six comparator architectures were evaluated: U-Net
\cite{ronneberger2015unet}, UNet++ \cite{zhou2018unetplusplus},
Attention U-Net \cite{oktay2018attentionunet}, DeepLabV3+
\cite{chen2018deeplabv3plus}, MobileNet-UNet, and a fetal-domain
SAM-based implementation denoted FetSAM. All comparators were trained
on the identical split with the same augmentation pipeline, optimiser,
and hardware.

Segmentation performance was evaluated using Dice coefficient,
Intersection over Union (IoU), precision, and recall for overlap;
95th-percentile Hausdorff distance (HD95) and average surface
distance (ASD) for boundary geometry; and atrial width mean absolute
error (Width MAE) for clinical measurement. Computational efficiency
was assessed using parameter count, model size, GFLOPs, and frames
per second (FPS) at batch size 1 on an NVIDIA Tesla T4 GPU. All
metrics except FPS and GFLOPs are reported as mean $\pm$ standard
deviation over five independent runs.

\subsection*{3.2. Overlap and Detection Performance}

Quantitative results are reported in Tables~1, 2, and 3.

\begin{table}[H]
\centering
\small
\setlength{\tabcolsep}{4pt}
\caption{Overlap and detection performance on the held-out test set
(mean $\pm$ std over five runs). Higher is better for all metrics.
Best values in \textbf{bold}.}
\label{tab:segmentation_overlap}
\begin{tabular}{@{}lcccc@{}}
\toprule
Model & Dice (\%) & IoU (\%) & Precision (\%) & Recall (\%) \\
\midrule
U-Net
  & $60.45 \pm 1.44$ & $45.63 \pm 1.46$
  & $47.73 \pm 2.04$ & $93.17 \pm 0.76$ \\
UNet++
  & $72.01 \pm 2.36$ & $58.27 \pm 3.22$
  & $63.43 \pm 4.25$ & $91.53 \pm 2.80$ \\
Attention U-Net
  & $75.58 \pm 0.39$ & $63.39 \pm 0.27$
  & $74.18 \pm 1.41$ & $82.53 \pm 0.51$ \\
FetSAM
  & $69.81 \pm 15.11$ & $57.51 \pm 15.75$
  & $72.36 \pm 13.33$ & $72.54 \pm 15.91$ \\
DeepLabV3+
  & $79.51 \pm 0.38$ & $66.73 \pm 0.49$
  & $78.73 \pm 1.93$ & $85.59 \pm 1.69$ \\
MobileNet-UNet
  & $79.55 \pm 0.86$ & $67.32 \pm 0.94$
  & $78.93 \pm 1.61$ & $84.86 \pm 0.46$ \\
\midrule
MVSegNet (Proposed)
  & $\mathbf{80.79 \pm 0.54}$
  & $\mathbf{68.47 \pm 0.70}$
  & $\mathbf{81.37 \pm 0.89}$
  & $84.53 \pm 1.35$ \\
\bottomrule
\end{tabular}
\end{table}

MVSegNet achieves the highest Dice ($80.79 \pm 0.54$\%) and IoU
($68.47 \pm 0.70$\%) among all evaluated models. The margin over
MobileNet-UNet ($79.55 \pm 0.86$\%) and DeepLabV3+
($79.51 \pm 0.38$\%) is modest but consistent across five runs.
U-Net ($60.45 \pm 1.44$\%) and UNet++ ($72.01 \pm 2.36$\%) show
that heavier generic architectures do not outperform compact
task-specific designs on this dataset.

MVSegNet precision ($81.37 \pm 0.89$\%) exceeds all comparators.
Recall ($84.53 \pm 1.35$\%) is marginally below DeepLabV3+ and
MobileNet-UNet. This trade-off is intentional: downstream measurement
depends on boundary placement, so higher precision with a small
reduction in recall is the preferred operating point.

\subsection*{3.3. Boundary and Measurement Performance}

\begin{table}[H]
\centering
\caption{Boundary and clinical measurement performance on the
held-out test set (mean $\pm$ std over five runs). Lower is better
for all metrics. Best values in \textbf{bold}.}
\label{tab:segmentation_boundary}
\begin{tabular}{lccc}
\toprule
Model & HD95 (mm) & ASD (mm) & Width MAE (mm) \\
\midrule
U-Net
  & $41.06 \pm 8.56$ & $7.69 \pm 1.43$ & $33.62 \pm 5.74$ \\
UNet++
  & $22.34 \pm 5.77$ & $4.09 \pm 0.92$ & $20.22 \pm 5.19$ \\
Attention U-Net
  & $7.70  \pm 2.52$ & $1.62 \pm 0.40$ & $7.45  \pm 1.32$ \\
FetSAM
  & $4.63  \pm 0.28$ & $1.06 \pm 0.05$ & $4.82  \pm 0.54$ \\
MobileNet-UNet
  & $4.90  \pm 1.16$ & $1.18 \pm 0.17$ & $4.27  \pm 1.09$ \\
DeepLabV3+
  & $4.28  \pm 0.09$ & $1.03 \pm 0.03$ & $4.12  \pm 0.33$ \\
\midrule
MVSegNet (Proposed)
  & $\mathbf{4.07 \pm 0.23}$
  & $\mathbf{0.98 \pm 0.05}$
  & $\mathbf{3.40 \pm 0.14}$ \\
\bottomrule
\end{tabular}
\end{table}

MVSegNet achieves the best values on all three boundary metrics:
HD95 ($4.07 \pm 0.23$~mm), ASD ($0.98 \pm 0.05$~mm), and Width MAE
($3.40 \pm 0.14$~mm). The standard deviations of HD95
($\pm 0.23$~mm) and Width MAE ($\pm 0.14$~mm) are the lowest among
all models, showing that gains are consistent across runs rather
than the result of a single favourable initialisation.

U-Net produces a Width MAE of $33.62 \pm 5.74$~mm, nearly ten times
that of MVSegNet, confirming that generic overlap-trained models are
not suitable for measurement-oriented ventricle analysis. UNet++ and
Attention U-Net close this gap but remain well above MVSegNet while
running at four to forty times the computational cost.

\subsection*{3.4. Computational Efficiency}

\begin{table}[H]
\centering
\caption{Computational complexity and inference throughput. GFLOPs
and parameter counts are for a single $512 \times 512$ input. FPS
is measured on an NVIDIA Tesla T4 GPU at batch size 1. Best values
in \textbf{bold}.}
\label{tab:efficiency}
\begin{tabular}{lrrrr}
\toprule
Model & Params (M) & Size (MB) & GFLOPs & FPS \\
\midrule
U-Net           & 7.43 & 28.30 & 106.33 & 27.6  \\
UNet++          & 2.17 & 8.28  & 64.59  & 25.6  \\
Attention U-Net & 7.56 & 28.83 & 107.96 & 24.3  \\
FetSAM          & 7.90 & 30.10 & 8.57   & 87.5  \\
MobileNet-UNet  & 4.27 & 16.27 & 12.20  & 83.3  \\
DeepLabV3+      & 1.42 & 5.43  & 16.40  & 98.9  \\
\midrule
MVSegNet (Proposed) & 2.31 & 8.80
  & \textbf{2.92} & \textbf{165.6} \\
\bottomrule
\end{tabular}
\end{table}

U-Net and Attention U-Net exceed 106~GFLOPs with throughput below
28~FPS. UNet++ incurs 64.59~GFLOPs at 25.6~FPS. Among the efficient
models, MVSegNet records the lowest GFLOPs (2.92) and the highest
throughput (165.6~FPS), a 1.67$\times$ speedup over DeepLabV3+
(98.9~FPS), while surpassing it on every accuracy and boundary
metric. The model footprint of 8.80~MB supports use in
memory-constrained settings.

\subsection*{3.5. Qualitative Segmentation Results}

Fig.~2 shows segmentation outputs for a representative
transventricular frame across all models. MVSegNet produces a
prediction that closely follows the annotated atrial region. Several
baselines show boundary leakage, fragmented predictions, or
over-segmentation. U-Net and UNet++ show the most extensive
false-positive regions, consistent with their weaker HD95 and Width
MAE values.

\begin{figure}[H]
    \centering
    \includegraphics[width=\textwidth,height=0.85\textheight,
    keepaspectratio]{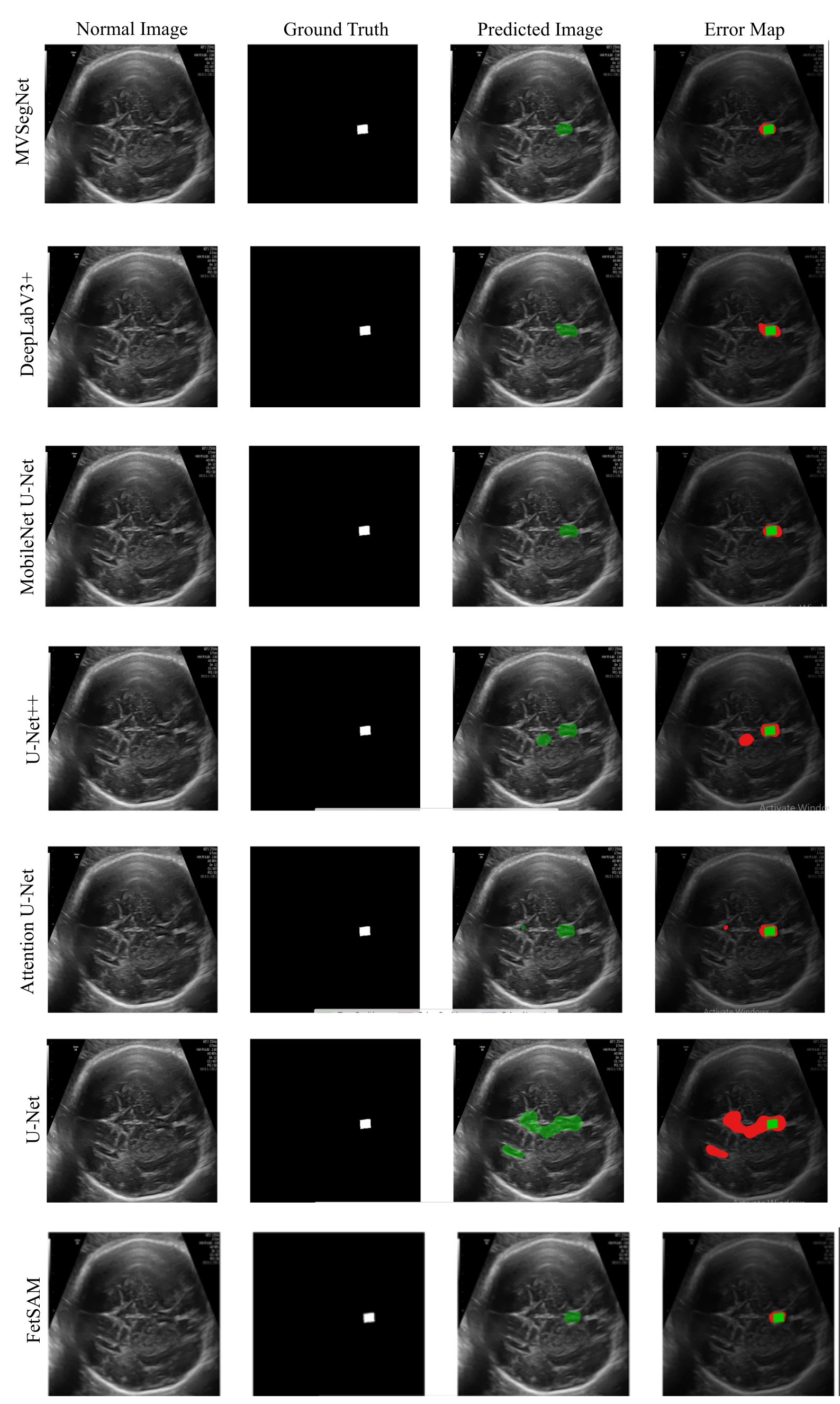}
    \caption{Qualitative comparison across architectures for a
    representative transventricular frame. Rows (top to bottom):
    MVSegNet, MobileNet-UNet, Attention U-Net, UNet++, DeepLabV3+,
    U-Net, FetSAM. Columns (left to right): input frame,
    ground-truth annotation, predicted mask (green overlay), error
    map. MVSegNet shows better boundary fidelity. U-Net and UNet++
    show over-segmentation in acoustically difficult regions.}
    \label{fig:qualitative}
\end{figure}
\FloatBarrier

\subsection*{3.6. Ablation Study}

To examine component contributions, we conducted an incremental
ablation study adding modules progressively on top of a shared
MobileNetV3-Small encoder-decoder backbone. Six configurations were
evaluated under the same conditions as the main comparison. Results
are mean $\pm$ standard deviation over three independent runs and
are summarised in Table~4.

\begin{table}[H]
\centering
\caption{Incremental ablation study. \textbf{Y} indicates the
component is active. AG: attention gates; MSAM: Multi-Scale
Ventricle Attention Module; FiLM: gestational-age conditioning;
ABRB: Adaptive Boundary Refinement Block; DS: deep supervision.
Metrics are mean $\pm$ std over three runs. Lower is better for
HD95, ASD, and Width MAE; higher is better for Dice.}
\label{tab:ablation}
\resizebox{\textwidth}{!}{%
\begin{tabular}{@{}llccccc cccc@{}}
\toprule
\multirow{2}{*}{Config.} & \multirow{2}{*}{Description}
  & \multicolumn{5}{c}{Components}
  & \multicolumn{4}{c}{Metrics} \\
\cmidrule(lr){3-7}\cmidrule(lr){8-11}
  & & AG & MSAM & FiLM & ABRB & DS
    & Dice (\%) & HD95 (mm) & ASD (mm) & W-MAE (mm) \\
\midrule
V1 & Base
   & & & & &
   & $79.84 \pm 0.54$ & $4.18 \pm 0.13$
   & $1.01 \pm 0.02$ & $3.77 \pm 0.29$ \\
V2 & +AG
   & Y & & & &
   & $79.62 \pm 0.77$ & $4.30 \pm 0.28$
   & $1.01 \pm 0.05$ & $3.94 \pm 0.36$ \\
V3 & +MSAM
   & Y & Y & & &
   & $79.64 \pm 0.75$ & $4.90 \pm 1.01$
   & $1.13 \pm 0.16$ & $4.67 \pm 1.23$ \\
V4 & +FiLM
   & Y & Y & Y & &
   & $79.36 \pm 0.74$ & $4.21 \pm 0.22$
   & $1.01 \pm 0.04$ & $3.84 \pm 0.08$ \\
V5 & +ABRB
   & Y & Y & Y & Y &
   & $79.44 \pm 0.45$ & $4.50 \pm 0.56$
   & $1.08 \pm 0.11$ & $4.07 \pm 0.54$ \\
V6 & +DS (Full)
   & Y & Y & Y & Y & Y
   & $\mathbf{80.79 \pm 0.54}$
   & $\mathbf{4.07 \pm 0.23}$
   & $\mathbf{0.98 \pm 0.05}$
   & $\mathbf{3.40 \pm 0.14}$ \\
\bottomrule
\end{tabular}}
\end{table}

Adding attention gates (V1$\rightarrow$V2) produced marginal changes
in all metrics. Adding MSAM (V2$\rightarrow$V3) left mean Dice
unchanged but increased variance in HD95 and Width MAE. Introducing
FiLM (V3$\rightarrow$V4) reduced HD95 standard deviation from
$\pm 1.01$~mm to $\pm 0.22$~mm and Width MAE standard deviation
from $\pm 1.23$~mm to $\pm 0.08$~mm, showing that gestational age
conditioning stabilises boundary predictions. Adding ABRB
(V4$\rightarrow$V5) further reduced Dice variance. The full model
(V6) achieved the best values on all four metrics, confirming that
all components are needed together.

\section*{4. Discussion}

\subsection*{4.1. Performance}

MVSegNet achieved the best overall balance of overlap accuracy,
boundary quality, and computational cost among the evaluated models.
The best HD95 and ASD values indicate reliable contour localisation
in difficult regions. The average surface distance below 1~mm shows
that predicted contours remain close to expert annotations on
average.

The width estimation results should be read with caution. The lowest
Width MAE among the compared methods suggests that better
segmentation supports more consistent measurement, but the remaining
error means the model should be viewed as a support tool rather than
a replacement for direct clinical measurement.

\subsection*{4.2. Model Design and Performance}

The proposed model performed better than heavier general-purpose
baselines on this dataset. Fetal lateral ventricle ultrasound has a
narrow appearance space where local boundary quality matters more
than broad visual diversity. Multi-scale feature extraction captures
both local detail and broader context, while boundary-aware
refinement supports more precise contour recovery. The efficient
backbone keeps the model compact and fast. Careful design choices
appear more useful here than increasing model size.

\subsection*{4.3. Efficiency}

MVSegNet recorded the lowest GFLOPs and the highest inference speed
while also giving the best segmentation results. The small model
size and low computational cost make it a practical candidate for
settings with limited hardware, though device-specific profiling
and regulatory validation would be needed before clinical use.

\subsection*{4.4. Limitations}

\textbf{Single dataset.} All experiments used 584 images from one
public dataset. Validation on external datasets from different
centres and scanners is needed to assess generalisation.

\textbf{Image-level split.} The dataset provides no patient-level
identifiers, so the split was done at image level. Results should
be read as benchmark performance rather than a subject-level
generalisation estimate.

\textbf{Single partition.} Results are over one fixed split with
multiple random seeds. Cross-validation across multiple partitions
would give a broader view of stability.

\textbf{Width estimation method.} Atrial width was estimated from
the maximum horizontal mask extent, which does not reproduce the
clinical inner-to-inner caliper protocol. A more faithful
measurement method is needed before clinical substitution.

\textbf{Hardware scope.} Inference speed was measured on an NVIDIA
T4 GPU. Runtime will differ on embedded hardware or clinical
ultrasound platforms.

\section*{5. Conclusions}

This paper presented MVSegNet, a lightweight network for fetal
lateral ventricle segmentation in prenatal ultrasound. The model
was designed to improve segmentation quality and boundary
localisation while keeping computational cost low.

On the public transventricular ultrasound benchmark, MVSegNet
achieved the best overall results among the evaluated methods on
Dice, IoU, HD95, ASD, GFLOPs, and inference speed. Improved
segmentation also supported more consistent atrial width estimation,
though this remains secondary to the main segmentation objective.

A compact, carefully designed model can perform well on this task
without high computational cost. Future work should address broader
dataset validation, evaluation across multiple data partitions, and
a more clinically faithful width estimation method.

\section*{CRediT Author Contribution Statement}

\textbf{Arafat Hossain Sayem:} Conceptualisation, Methodology,
Software, Formal analysis, Investigation, Data curation,
Visualisation, Writing -- original draft, Writing -- review and
editing.

\section*{Declaration of Competing Interest}

The authors declare that they have no known competing financial
interests or personal relationships that could have appeared to
influence the work reported in this paper.

\section*{Data Availability}

The dataset used in this study is publicly available. The
transventricular ultrasound dataset was released by
\cite{alzubaidi2023dataset} and can be accessed at the repository
provided in the original publication. Model code and training
scripts will be made available upon acceptance.

\section*{Funding}

This research received no specific grant from any funding agency
in the public, commercial, or not-for-profit sectors. Computational
experiments were conducted using an NVIDIA Tesla T4 GPU provided
through Google Colaboratory (Google LLC, Mountain View, CA, USA).

\section*{Ethics Statement}

This study used a publicly available, fully de-identified dataset
\cite{alzubaidi2023dataset} in which no patient identifiers were
retained. All data were collected and released by the original
authors in accordance with applicable institutional and ethical
guidelines. No direct interaction with human participants occurred
and no identifiable personal data were accessed, so formal ethical
approval was not required. The research followed the ethical
principles of the Declaration of Helsinki.


\end{document}